\documentclass[letterpaper, 10 pt, conference]{ieeeconf}  

\IEEEoverridecommandlockouts                              

\usepackage{graphics} 
\usepackage{epsfig} 
\usepackage{times} 
\usepackage{amsmath} 
\usepackage{amssymb}  
\usepackage{graphicx}
\usepackage{hyperref}
\hypersetup{hidelinks}
\usepackage{xcolor}
\usepackage{makecell}
\usepackage{physics}
\usepackage{mathtools}
\usepackage{float}

\usepackage{booktabs}
\usepackage{wrapfig}
\usepackage{tabu}
\usepackage[ruled,vlined,algo2e]{algorithm2e}
\usepackage{algorithm}
\usepackage{algorithmic}
\usepackage{multirow}
\let\labelindent\relax
\usepackage{enumitem}

\usepackage[caption=false,font=footnotesize]{subfig}

\newcommand{\bfsection}[1]{\vspace*{0.1cm}\noindent\textbf{#1. }}

\makeatletter
\def\endthebibliography{%
  \def\@noitemerr{\@latex@warning{Empty `thebibliography' environment}}%
  \endlist
}
\makeatother

\usepackage{xspace}
\makeatletter
\DeclareRobustCommand\onedot{\futurelet\@let@token\@onedot}
\def\@onedot{\ifx\@let@token.\else.\null\fi\xspace}

\def\ie{\emph{i.e}\onedot}

\makeatother

\title{\LARGE \bf A Strong View-Free Baseline Approach for Single-View Image Guided Point Cloud Completion}

\author{Fangzhou Lin$^{*1,3}$, Zilin Dai$^{*5}$, Rigved Sanku$^{*1}$, Songlin Hou$^{2,4}$, \\Kazunori D Yamada$^{3}$, Haichong K. Zhang$^{1,4,6}$, and Ziming Zhang$^{\dag,1,6,7}$  
\thanks{*These authors contributed equally to this work.}
\thanks{\dag Corresponding author.}
\thanks{$^{1}$Department of Robotics Engineering, Worcester Polytechnic Institute, Worcester, MA 01609, USA}%
\thanks{$^{2}$Dell Technologies, Hopkinton, MA 01748, USA}%
\thanks{$^{3}$Graduate School of Information Sciences, Tohoku University, Sendai, 980-8579, Japan}
\thanks{$^{4}$Department of Computer Science, Worcester Polytechnic Institute, Worcester, MA 01609, USA}
\thanks{$^{5}$Harvard Kenneth C. Griffin Graduate School of Arts and Sciences, Cambridge, MA 02138, USA}
\thanks{$^{6}$Department of Biomedical Engineering, Worcester Polytechnic Institute, Worcester, MA 01609, USA}
\thanks{$^{7}$Department of Electrical \& Computer Engineering, Worcester Polytechnic Institute, Worcester, MA 01609, USA 
({\tt\small zzhang15@wpi.edu})}
}
\begin{document}

\maketitle
\thispagestyle{empty}
\pagestyle{empty}

\begin{abstract}

The {\em single-view} image guided point cloud completion (SVIPC) task aims to reconstruct a complete point cloud from a partial input with the help of a single-view image. While previous works have demonstrated the effectiveness of this multimodal approach, the fundamental necessity of image guidance remains largely unexamined. To explore this, we propose a strong baseline approach for SVIPC based on an attention-based multi-branch encoder-decoder network that only takes partial point clouds as input, \ie view-free. Our hierarchical self-fusion mechanism, driven by cross-attention and self-attention layers, effectively integrates information across multiple streams, enriching feature representations and strengthening the network’s ability to capture geometric structures. Extensive experiments and ablation studies on the ShapeNet-ViPC dataset demonstrate that our view-free framework performs superiorly to state-of-the-art SVIPC methods. We hope our findings provide new insights into the development of multimodal learning in SVIPC. {\it Our demo code will be available at \url{https://github.com/Zhang-VISLab}.}

\end{abstract}

\section{Introduction}

\bfsection{Background}
Recent advancements in 3D perception have significantly increased research and industrial interest in point cloud completion, a crucial task for robotics, autonomous driving, and augmented reality \cite{xu2022fpcc}\cite{hou2025mobile}. However, real-world sensor data often contains incomplete point clouds due to factors such as occlusions, restricted viewpoints, and noise \cite{lin2022cosmos}. To address these issues, deep learning–based methods have substantially enhanced reconstruction quality by understanding complex spatial relationships and learning how to map sparse inputs to dense 3D shapes \cite{wang2019dynamic}\cite{zhang2021point}, paving the way for more advanced network architectures that achieve state-of-the-art performance in point cloud completion.

Existing research on deep learning for point cloud completion has predominantly utilized an encoder-decoder architecture, where the encoder transforms incomplete input into a latent feature space that captures both global structure and local geometric details, while the decoder reconstructs a dense 3D representation from these features \cite{yuan2018pcn}. Building on this foundation, methods such as AtlasNet \cite{AtlasNet}, FoldingNet \cite{yang2018foldingnet}, and TopNet \cite{tchapmi2019topnet} have introduced alternative decoder architectures designed to enhance the representation of surface geometries, leading to sharper and more coherent reconstructions. More recently, several studies have explored complementary techniques, including attention mechanisms, to refine the modeling of point relationships and improve inference performance \cite{PoinTr}. However, the quality of the latent representation in the encoder-decoder framework, as commonly found in existing works, can be a limiting factor in enhancing generation performance. Constructing a latent representation with rich information, while often non-trivial, is crucial for effective point cloud completion.
\newline
\indent To construct a latent representation with richer information, some researches adopt multi-modal design to fuse inputs from multiple sources, of which a popular design is view-based methods. A popular approach within this design is view-based methods, which incorporate additional sources of information to enhance point cloud representations. Contrary to traditional deep learning methods which only take incomplete point clouds as inputs, view-based point cloud completion methods, typically require images as additional input alongside the incomplete point cloud. Single-view images provide complementary geometric cues, helping the model to infer missing structures that may be unclear from partial 3D data alone. These methods leverage image features to address occlusions, maintain viewpoint consistency, and often improve completion accuracy by guiding the network toward a more informed reconstruction. For instance, ViPC \cite{zhang2021view} generates a coarse point cloud from a single-view image before refinement and employs vector concatenation to fuse the two modalities. DMF-Net \cite{lu2024mmcnet} adopts a dual-channel approach, with one channel processing the partial point cloud and the other processing an image of the object, placing greater emphasis on fusion balance through a symmetric design. CSDN \cite{zhu2023csdn} utilizes a shape-transfer mechanism to extract shape characteristics from the input image, guiding the completion process. The image used in CSDN input, which contains the missing parts of the point cloud, assists the network in identifying the regions requiring completion. Song \textit{et al.} \cite{song2023fine} leverages a pretrained CLIP \cite{radford2021learning} model in a multimodal fusion setting to incorporate both visual and textual information for guiding point cloud completion. In these methods, images, serving as 2D projections of 3D objects, provide additional information about the objects' appearance and geometry to facilitate the completion process. 

\bfsection{Challenges in SVIPC}
While incorporating images alone may enhance the completion process by providing additional information, it also introduces additional concerns and limitations. For instance, the performance of view-based methods heavily depends on the quality of input images and the chosen viewpoint, and blurry, occluded, or unfavorably angled images can negatively impact performance \cite{fu2023vapcnet}. Additional efforts are required for camera calibration, as misalignment or calibration errors can degrade completion performance \cite{tang2025calibration}. Subsequently, images alone may not provide sufficient geometrical information in certain scenarios, especially for 3D shapes lacking texture or containing ambiguous regions \cite{zhou2024position}. 
The process of capturing and processing image inputs introduces significant computational complexity, which raises concerns regarding the scalability of SVIPC in real-world applications. These challenges often become even more pronounced, where obtaining a high-quality image specifically associated with the partial point cloud may prove difficult or even infeasible. Surprisingly, there are few studies in the literature that explore the necessity of single-view image guidance in point cloud completion tasks, which significantly motivates our work.

\bfsection{Our Approach \& Contributions}
To explore this necessity, we propose a strong view-free baseline method for SVIPC. We introduce a multi-branch 3D encoder network that operates solely on partial point clouds. The 3D encoder, built upon PointNet++ \cite{qi2017pointnet++}, extracts multi-scale geometric features and captures both fine-grained local details and global structures. These features are then progressively refined through self-fusion, which leverages multi-stage attentions to guide the completion process by fusing information within the partial point clouds. We term this approach ``self-fusion,'' as it enriches feature representations by integrating geometric cues directly from the input data. This hierarchical fusion mechanism ensures that the model dynamically adjusts its focus, strengthening representations at different abstraction levels. The multi-branch design enables diverse feature learning, where different stages specialize in capturing different geometric aspects. By integrating these complementary representations, our approach improves reconstruction robustness without relying on external view cues.

We conducted extensive experiments and ablation studies on the ShapeNet-ViPC dataset to validate our approach. Our results demonstrate that the proposed view-free framework can achieve superior performance compared to state-of-the-art SVIPC methods. These findings suggest that view-free approaches can be highly effective for point cloud completion. We hope our work provides new insights into the development in SVIPC and encourages further exploration of view-free techniques for point cloud completion.

\section{Related Work}
\label{sec:rw}

Point cloud completion has been extensively studied in 3D computer vision due to its significance in applications such as autonomous driving, robotics, and scene reconstruction \cite{xu2022fpcc, lin2022cosmos,zhang2024deep,gao2025safecoop}. Early approaches to shape completion relied on explicit geometric descriptors or shape retrieval from large datasets, with methods such as data-driven structural priors \cite{Data-Driven,yue2024understanding} and primitive shape-based reconstructions \cite{veryold}, though they often struggled with generalizing to complex and diverse geometries. With the advent of deep learning, encoder-decoder architectures such as PCN \cite{PCN} have become prominent, as they eliminate assumptions about structural priors. Researchers have since proposed a variety of architectures and loss functions \cite{lin2024loss,lin2023infocd,lin2023hyperbolic,lin2024hyperbolic,zhang2025gps} to further optimize point cloud completion performance. Subsequent works, including AtlasNet \cite{AtlasNet} and TopNet \cite{TopNet}, advanced point cloud generation by leveraging parametric surface elements and hierarchical structures.

Transformer-based methods have also gained traction. For example, models like PoinTr \cite{PoinTr} employ geometry-aware transformers to capture local relationships in 3D spaces, while SnowflakeNet \cite{SnowflakeNet} refines point distributions through iterative splitting to produce highly detailed reconstructions. Recent advancements such as VRCNet \cite{Variational} incorporate dual-path architectures for probabilistic modeling, enhancing both accuracy and robustness.

The integration of auxiliary 2D image data for point cloud completion represents a new direction. Initiated by ViPC \cite{zhang2021view}, view-guided completion introduced the use of single-view reconstructions as priors. However, its reliance on an intermediate coarse point cloud generation can limit performance. Building on this foundation, subsequent works have explored more advanced fusion strategies; for instance, XMFNet \cite{XMFNet} leverages a Dynamic Graph CNN for point cloud feature extraction and integrates multiple cross and self-attention layers to achieve single-level feature fusion between image and point cloud modalities, establishing a strong baseline for multimodal 3D reconstruction. More recently, additional view-guided methods have emerged. For example, CSDN \cite{zhu2023csdn} leverages shape-transfer to propagate geometric details from global shape to point cloud domain and dual-refinement strategies to iteratively enhance global shape structures and local details; Fine-grained CLIP \cite{song2023fine} exploits textual cues in combination with images for more contextually aware shape completion; DMF-Net \cite{mao2024dmf} and EGIInet \cite{xu2024explicitly} introduce dual-channel modality fusion and shape-aware upsampling to ensure finer geometric consistency for further performance improvement.

\section{Our Approach}
\label{sec:approach}

\begin{figure*}[htb!]
    \centering
    \includegraphics[width=0.8\linewidth]{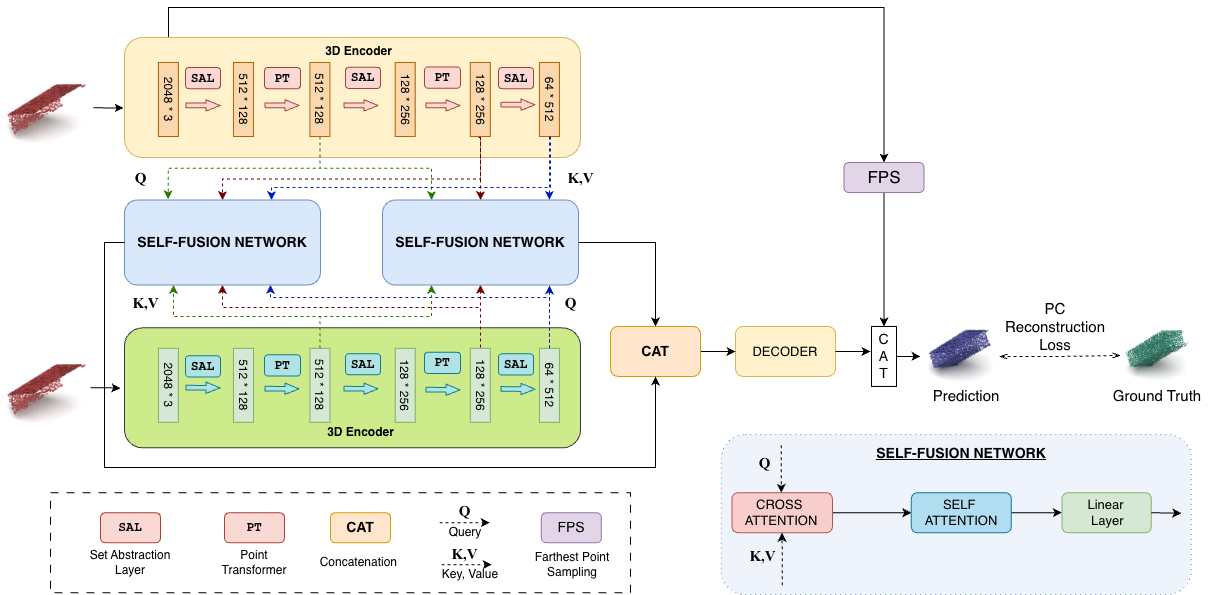}
    \caption{Architecture of our attention-enhanced self-fusion network with two branches. (We used three branches for performance comparison in experiments, which can be easily extended from this two-branch case) The incomplete point cloud is processed by two 3D encoders, each extracting hierarchical features across three levels using Set Abstraction Layers (SAL) and Point Transformers (PT). Intermediate features from all levels are fed into the self-fusion network, where cross-attention and self-attention refine the representations. The fused features (six in total—three from each encoder coming from 2 self-fusion networks) are concatenated and passed through a decoder. The decoder, along with the original point cloud, generates a refined point cloud. Farthest Point Sampling (FPS) ensures uniform point distribution, while reconstruction loss guides the learning process.}
    \label{fig:net_structure}
    \vspace{-0.3cm}
\end{figure*}

Our proposed method utilizes only point cloud data as input, in contrast to recent SVIPC networks that incorporate both point clouds and images. The network comprises of two (or more) branches, each receiving the point cloud as input. Features extracted from each branch undergo self-fusion, where they are progressively refined before being aggregated and passed to the decoder for the point cloud reconstruction. For simplicity, we illustrate a two-branch version of our network structure in Figure \ref{fig:net_structure}.


\bfsection{3D Encoder} 
Taking an incomplete point cloud as input, the initial features are extracted using a 3D encoder. Our 3D encoder follows a modular design, allowing the use of various feature extractors such as PointNet++ \cite{qi2017pointnet++} and DGCNN \cite{wang2019dynamic}. Inspired by SnowflakeNet, our 3D encoder adopts a Set Abstraction (SA) layer with K-NN. Unlike SnowflakeNet, which retains only the final feature, our encoder retains both intermediate feature representations and the final encoded feature, as shown in the orange and green boxes in Figure \ref{fig:net_structure}. In contrast to the single-feature extraction in SnowflakeNet and XMFNet, our 3D encoder extracts hierarchical features at multiple levels, capturing both global structural information and local fine-grained details from the input point cloud. In the two-branch configuration, each branch processes the input point cloud independently through the 3D encoder, ensuring diverse feature extraction. These multi-scale representations are then grouped by hierarchical level before being processed in the self-fusion network.

\bfsection{Self-Fusion Network}
The self-fusion network refines the extracted features through a combination of cross-attention, self-attention, and a linear projection layer. Both cross-attention and self-attention follow a multi-head attention design. In cross-attention, features from different branches at the same hierarchical level interact, with one serving as the query (\( Q \)) and the other as the key (\( K \)) and value (\( V \)). In self-attention, each feature attends to itself to enhance its contextual representation. Residual connections and layer-wise normalization are applied after each attention layer, ensuring stable feature refinement. The linear layer projects feature dimensions up to 512, allowing feature concatenation across different levels.

Stacking cross-attention and self-attention for each branch enables the network to iteratively blend fused features while refining them through augmented context. Unlike simple fusion strategies (e.g., concatenation or element-wise addition), this attention-based fusion adaptively balances feature contributions across branches, leading to richer and highly adaptive, context-enriched latent representations.

\bfsection{Architecture with Three and Four Branches} 
We design a unified architecture that can scale flexibly to multiple branches. We evaluate its efficacy on two, three, and four-branch configurations to assess performance variations. In all cases, each branch has its own feature encoder and produces a set of latent tokens. Positional encodings are added to each branch's tokens to retain spatial structure, aiding the self-fusion network in learning feature correspondences across branches. The branches are then fused via a series of self-fusion networks, where features from each pair of branches is blended using two self-fusion networks to facilitate effective feature interaction. Each self-fusion network follow the same structure as described in Fig \ref{fig:net_structure}.

\bfsection{Point Cloud Generation Decoder}
The decoder reconstructs the complete point cloud by estimating the missing points and integrating them with the input partial cloud. Since decoding the acquired fusion features requires flexibility and a strong learning capacity, we adopt a decoder architecture similar to XMFNet \cite{wang2022pointattn}, designed to process features and learn their implicit shape representations to predict 3D coordinates. Our plug-and-play attention-enhanced self-fusion network is compatible with various other point cloud generation decoders, such as one used by PointAttN \cite{wang2022pointattn}.

\bfsection{Loss Function} 
As it is a supervised learning task, Chamfer Distance ($L_{CD}$) is used to minimize the discrepancy between the predicted ($\hat{Y}$) and ground truth ($Y$) point clouds:
\[
L_{CD}(Y, \hat{Y}) = \frac{1}{|Y|} \sum_{y \in Y} \min_{\hat{y} \in \hat{Y}} \|y - \hat{y}\|_2^2 + \frac{1}{|\hat{Y}|} \sum_{\hat{y} \in \hat{Y}} \min_{y \in Y} \|y - \hat{y}\|_2^2.
\]



\section{Experiments and Results}

\subsection{Datasets}
All experiments are conducted on the ShapeNet-ViPC dataset, which contains 38,328 objects from 13 categories. Each object is associated with 24 partial point clouds and corresponding images, generated under 24 viewpoints, following the previous works' standard setting \cite{zhang2021view,zhu2023csdn,XMFNet}. 

\subsection{Evaluation Metrics}
Following the literature, the performance is evaluated using Chamfer Distance (CD) and F-Score. CD measures the average squared distance between points in the predicted and ground truth point clouds, assessing reconstruction accuracy. F-Score computes overlap precision and recall at a fixed threshold, reflecting structural similarity \cite{zhang2021view,xu2024explicitly}.

\begin{table}[h]
  \begin{center}
  \setlength{\tabcolsep}{1pt}{\footnotesize
  \caption{Mean CD per point ($\text{CD}\times 10^3\downarrow$) on ShapeNet-ViPC}
  \label{table:1}
    \begin{tabular}{l|c c c c c c c c c}
    \toprule
    Methods&Avg&Airplane&Cabinet&Car&Chair&Lamp&Sofa&Table&Watercraft \\
    \midrule
    AtlasNet&6.062&5.032&6.414&4.868&8.161&7.182&6.023&6.561&4.261 \\
    FoldingNet&6.271&5.242&6.958&5.307&8.823&6.504&6.368&7.080&3.882 \\
    PCN&5.619&4.246&6.409&4.840&7.441&6.331&5.668&6.508&3.510 \\
    TopNet&4.976&3.710&5.629&4.530&6.391&5.547&5.281&5.381&3.350 \\
    PF-Net&3.873&2.515&4.453&3.602&4.478&5.185&4.113&3.838&2.871 \\
    MSN&3.793&2.038&5.060&4.322&4.135&4.247&4.183&3.976&2.379 \\
    GRNet&3.171&1.916&4.468&3.915&3.402&3.034&3.872&3.071&2.160 \\
    PoinTr&2.851&1.686&4.001&3.203&3.111&2.928&3.507&2.845&1.737 \\
    PointAttN&2.853&1.613&3.969&3.257&3.157&3.058&3.406&2.787&1.872 \\
    SDT&4.246&3.166&4.807&3.607&5.056&6.101&4.525&3.995&2.856 \\
    Seedformer&2.902&1.716&4.049&3.392&3.151&3.226&3.603&2.803&1.679 \\
    \midrule
    ViPC&3.308&1.760&4.558&3.183&2.476&2.867&4.481&4.990&2.197 \\
    CSDN&2.570&1.251&3.670&2.977&2.835&2.554&3.240&2.575&1.742 \\
    XMFnet&1.443&0.572&1.980&1.754&1.403&1.810&1.702&1.386&0.945 \\
    
    EGIInet&1.211&0.534&1.921&1.655&1.204&0.776&1.552&1.227&0.802 \\
    \midrule
    \textbf{Ours} & \textbf{1.102} & \textbf{0.490} & \textbf{1.742} & \textbf{1.549} & \textbf{1.198} & \textbf{0.670} &
    \textbf{1.471} &
    \textbf{1.190} &
    \textbf{0.762} 
    \\    
    \bottomrule
    \end{tabular} \vspace{-.5cm}
}
  
  \end{center}  
\end{table}

\begin{table}[h]
  \begin{center}
  \setlength{\tabcolsep}{1pt}{\footnotesize
  \caption{Mean F-Score @ 0.001$\uparrow$ on ShapeNet-ViPC}
   \label{table:2}
    \begin{tabular}{l|c c c c c c c c c}
    \toprule
    Methods&Avg&Airplane&Cabinet&Car&Chair&Lamp&Sofa&Table&Watercraft \\
    \midrule
    AtlasNet&0.410&0.509&0.304&0.379&0.326&0.426&0.318&0.469&0.551 \\
    FoldingNet&0.331&0.432&0.237&0.300&0.204&0.360&0.249&0.351&0.518 \\
    PCN&0.407&0.578&0.270&0.331&0.323&0.456&0.293&0.431&0.577 \\
    TopNet&0.467&0.593&0.358&0.405&0.388&0.491&0.361&0.528&0.615 \\
    PF-Net&0.551&0.718&0.399&0.453&0.489&0.559&0.409&0.614&0.656 \\
    MSN&0.578&0.798&0.378&0.380&0.562&0.652&0.410&0.615&0.708 \\
    GRNet&0.601&0.767&0.426&0.446&0.575&0.694&0.450&0.639&0.704 \\
    PoinTr&0.683&0.842&0.516&0.545&0.662&0.742&0.547&0.723&0.780 \\
    PointAttN&0.662&0.841&0.483&0.515&0.638&0.729&0.512&0.699&0.774 \\
    SDT&0.473&0.636&0.291&0.363&0.398&0.442&0.307&0.574&0.602 \\
    Seedformer&0.688&0.835&0.551&0.544&0.668&0.777&0.555&0.716&0.786 \\
    \midrule
    ViPC&0.591&0.803&0.451&0.512&0.529&0.706&0.434&0.594&0.730 \\
    CSDN&0.695&0.862&0.548&0.560&0.669&0.761&0.557&0.729&0.782 \\
    XMFnet&0.796&0.961&0.662&0.691&0.809&0.792&0.723&0.830&0.901 \\
   
    EGIInet&0.836&0.969&0.693& 0.723& 0.847& 0.919& 0.756& 0.857& 0.927 \\
    \midrule
     \textbf{Ours}                             & \textbf{0.851} & \textbf{0.973} & \textbf{0.721} & \textbf{0.754} & \textbf{0.882} & \textbf{0.934} 
     & \textbf{0.761} 
     & \textbf{0.862} 
     & \textbf{0.931} \\   
    \bottomrule
    \end{tabular}
    }
  \end{center}
  \vspace{-0.4cm}
\end{table}

\begin{figure}
  \centering
  \includegraphics[width=0.4\textwidth]{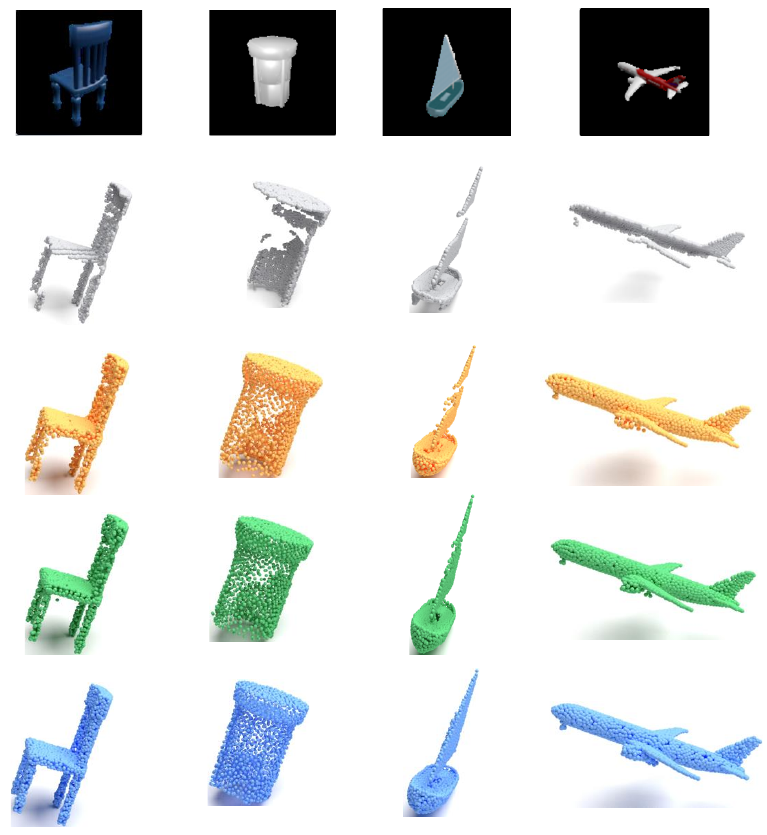}
  \caption{{\textbf{Row-1}}: Input images. {\textbf{Row-2}}: Incomplete point clouds. {\textbf{Row-3}}: View-guided (XMFNet) outputs. {\textbf{Row-4}}: {\bf Our view-free outputs}. {\textbf{Row-5}}: Ground truth. }
  \label{visulizaion}
  \vspace{-0.6cm}
\end{figure}

\subsection{Performance}

Table~\ref{table:1} and Table~\ref{table:2} summarize our quantitative comparison results for ShapeNet-ViPC dataset, where we report the best officially reported performance of each method based on prior work \cite{xu2024explicitly}. In these comparisons, we evaluate our method alongside existing view-guided point cloud completion approaches, including ViPC \cite{zhang2021view}, CSDN \cite{zhu2023csdn}, XMFNet \cite{XMFNet} and EGIInet \cite{xu2024explicitly}. Additionally, we benchmark against general point cloud completion methods, such as AtlasNet \cite{groueix2018papier}, FoldingNet \cite{yang2017foldingnet}, PCN \cite{yuan2018pcn}, TopNet \cite{Tchapmi_2019_CVPR}, PF-Net \cite{Huang_2020_CVPR}, MSN \cite{liu2020morphing}, GRNet \cite{xie2020grnet}, PoinTr \cite{yu2021pointr}, PointAttN \cite{wang2022pointattn}, SDT \cite{zhang2022point} and Seedformer \cite{zhou2022seedformer}. Our solution consistently outperforms methods mentioned above in terms of both CD and F-score, demonstrating strong reconstruction accuracy across multiple categories. 

Figure~\ref{visulizaion} presents a qualitative comparison, demonstrating that our view-free method produces reconstructions comparable to view-guided approaches. The visualization highlights our model's ability to generate structurally coherent point clouds, reinforcing its effectiveness as a strong baseline for SVIPC.


\subsection{Analysis Experiments}

\begin{wraptable}{r}{.59\linewidth}
        \vspace{-10mm}
	\centering
	\small
	\caption{Ablation study with Different way of Fusion ($\text{CD}\times 10^3\downarrow$).}
        \begin{tabular}{c|ccc}
    		\toprule
    		Fusions & Cabinet \\
    		\midrule
    		XMFnet & 1.980 \\     
            EGIInet & 1.921 \\     
            \midrule
            Two-Branch (single) & 2.238 \\
            Two-Branch (double)  & 1.802 \\
            Three-Branch  & 1.742 \\ 
            {\bf Four-Branch } & {\bf 1.738} \\
    		\bottomrule
	    \end{tabular}

	\label{table:3}
 \vspace{-3mm}
\end{wraptable}

To assess the effectiveness and robustness of our multi-branch  self-fusion architecture, we conduct a series of analysis experiments using various of network architecture as well as the point cloud data. In these experiments, all branches share the same backbone encoder but receive inputs that are either transformed or augmented versions of the original point cloud. The different configurations in considered, along with their corresponding results, are presented in Table~\ref{table:3}. Two versions of the two-branch setup were evaluated: (1) a single fusion network with one-way fusion, and (2) a standard double self-fusion network, where fusion occurs bidirectionally, with double fusion performing significantly better due to better feature integration.


\textbf{Three-Branch Fusion: } For the three-branch configuration, each branch processes the original partial point cloud independently. This setup evaluates whether introducing additional representations enhances robustness. The attention module dynamically learns to integrate complementary information from the three branches, yielding noticeable performance gains over the two-branch setup. This configuration effectively balances performance and computational cost, making it as the \textit{default choice for our method.}

\textbf{Four-Branch Fusion:}
We further extend the configuration into four branches, with each branch processing the same original partial point cloud data. While the four-branch model achieves further improvements in performance, the gains become less significant relative to the increased computational cost. This observation highlights a trade-off between enhanced feature diversity and model efficiency, suggesting that beyond a certain point, additional branches likely yield diminishing returns.


\textbf{Augmented Fusion:} In this setting, one branch processes the original point cloud, while the another branch receives an augmented version either rotated by a random angle or perturbed with additive Gaussian noise. This experiment investigates the effect of such augmentations on the network’s ability to learn robust representations. However, the results indicate a decline in performance, suggesting that these transformations may introduce inconsistencies that hinder effective feature fusion.



\vspace{-0.1cm}
\subsection{Ablation Study}

To assess the impact of different point feature feature extractors and generation decoders on this task, we conduct ablation studies using different combinations of feature extractors (DGCNN \cite{wang2019dynamic} and PointNet++ \cite{qi2017pointnet++}) and generation decoders (XMFNet-based and Transformer-based \cite{wang2022pointattn}). As shown in Table~\ref{table:4}, the self-fusion network demonstrates strong compatibility with different feature encoders and point cloud generation decoders, highlighting its flexibility and robustness. This study further validates the generality of the proposed self-fusion network across different architectural configurations.

\begin{table}[t]
    \centering
    \caption{Ablation studies for the point cloud feature extractors and generation decoders($\text{CD}\times 10^3\downarrow$).}
    
    \vspace{-2.5mm}
    \begin{center}
        \small{
        \begin{tabular}{ccc|c}
            \makecell{Feature \\Extractor } & \makecell{Point Generation \\ Decoder} &  &  Cabinet \\
            \midrule
            DGCNN & XMFNet-based & & \footnotesize 1.813 \\
            PoinNet++ & Transformer-based & & \footnotesize 1.917 \\
            DGCNN & Transformer-based & & \footnotesize 1.854 \\
            PointNet++ & XMFNet-based & & \footnotesize \textbf{1.742} \\
        \end{tabular}
        }
    \end{center}
    \label{table:4}
    \vspace{-0.9cm}
\end{table}

As extensively studied, vanilla CD is vulnerable to outliers, leading to the development of several CD variants including DCD \cite{wu2021densityaware}, HyperCD \cite{lin2023hyperbolic}, InfoCD \cite{lin2023infocd} and LandauCD \cite{lin2024loss}. To assess the impact of these variants as loss functions, we conduct an ablation study using a two-branch configuration, with results summarized in Table~\ref{table:Shapenet-Part_analysis2}. Our findings suggest that, without hyperparameter tuning, these CD variants provide only marginal improvements. Given the substantial effort required for hyperparameter optimization and our primary focus on pipeline design, we adopt vanilla CD as the default loss function for simplicity and efficiency.

\begin{wraptable}{r}{.45\linewidth}
    \vspace{-4mm}
	\centering
	\small
	\caption{Ablation study with Different Loss Functions ($\text{CD}\times 10^3\downarrow$).}
        \begin{tabular}{c|ccc}
    		\toprule
    		Loss Function & Cabinet \\
    		\midrule
    		CD & 1.802 \\     
            DCD & 1.814 \\
            HyperCD & 1.808 \\        
            InfoCD & 1.783 \\
            \textbf{LandauCD} & \textbf{1.781} \\ 
    		\bottomrule
	    \end{tabular}

	\label{table:Shapenet-Part_analysis2}
 \vspace{-3mm}
\end{wraptable}

We also provide the detailed complexity analysis of our method with different branch configurations in Table~\ref{tab:FLOPs}. We report the number of parameters and theoretical computation cost (FLOPs) of our method with different branches. The results show that the three-branch configuration achieves competitive performance while maintaining relatively low parameter count and FLOPs compared to other methods in the table. This demonstrates that our approach effectively balances computational efficiency and performance, making it a well-optimized choice for point cloud completion.

\begin{table}[h]
\small
\caption{Complexity analysis with different branches and its performance on cabinet category ($\text{CD}\times 10^3\downarrow$).} 
\label{tab:FLOPs}
\centering
\setlength{\tabcolsep}{1.5mm}{
\begin{tabular}[\linewidth]{l | r r | r  }
\toprule
Models & Params & FLOPs & CD \\
\midrule

Two-Branch    &7.752 M &6.936 G & 1.802 \\
Three-Branch & 7.759 M & 8.072 G & 1.742 \\
Four-Branch    & 8.808 M & 15.180 G & 1.738  \\
\midrule
XMFNet  & 8.725 M & 7.726 G & 1.980  \\

\bottomrule[1pt]
\end{tabular}}
\vspace{-0.5cm}
\end{table}

\section{Conclusion}
In this work, we introduce a strong baseline for SVIPC through an attention-based multi-branch encoder-decoder network that operates solely on partial point clouds. Our view-free architecture leverages cross- and self-attention mechanisms to effectively integrate multiple partial point clouds, enriching feature representations, and capturing fine-grained geometric details. Extensive experiments on the ShapeNet-ViPC dataset demonstrate that our method presents a robust alternative for view-guided point cloud completion. We hope that our research offers fresh insights into the advancements in SVIPC and inspires further exploration of both view-free and view-guided methods for the 3D point cloud completion task.

\bfsection{Limitations}
Given our network design, multiple branches may learn similar representations, causing feature redundancy. This can limit feature diversity despite improved robustness and may lead to overfitting as the network relies on repetitive patterns rather than diverse geometric cues. Future work could explore more adaptive learning mechanisms as well as diversity-promoting losses to mitigate this effect.

\newpage
\tiny
\bibliographystyle{IEEEtran}
\bibliography{references}

@misc{XMFNet,
      title={Cross-modal Learning for Image-Guided Point Cloud Shape Completion}, 
      author={Emanuele Aiello and Diego Valsesia and Enrico Magli},
      year={2022},
      eprint={2209.09552},
      archivePrefix={arXiv},
      primaryClass={cs.CV}
}

@INPROCEEDINGS{PCN,
  author={Yuan, Wentao and Khot, Tejas and Held, David and Mertz, Christoph and Hebert, Martial},
  booktitle={2018 International Conference on 3D Vision (3DV)}, 
  title={PCN: Point Completion Network}, 
  year={2018},
  volume={},
  number={},
  pages={728-737},
  doi={10.1109/3DV.2018.00088}}

@article{AtlasNet,
  author       = {Thibault Groueix and
                  Matthew Fisher and
                  Vladimir G. Kim and
                  Bryan C. Russell and
                  Mathieu Aubry},
  title        = {AtlasNet: {A} Papier-M{\^{a}}ch{\'{e}} Approach to Learning 3D
                  Surface Generation},
  journal      = {CoRR},
  volume       = {abs/1802.05384},
  year         = {2018},
  eprinttype    = {arXiv},
  eprint       = {1802.05384},
  bibsource    = {dblp computer science bibliography, https://dblp.org}
}

@article{Data-Driven,
author = {Sung, Minhyuk and Kim, Vladimir and Angst, Roland and Guibas, Leonidas},
year = {2015},
month = {10},
pages = {1-11},
title = {Data-Driven Structural Priors for Shape Completion},
volume = {34},
journal = {ACM Transactions on Graphics},
doi = {10.1145/2816795.2818094}
}

@article{veryold,
journal = {Computer Graphics Forum},
title = {{Completion and Reconstruction with Primitive Shapes}},
author = {Schnabel, Ruwen and Degener, Patrick and Klein, Reinhard},
year = {2009},
publisher = {The Eurographics Association and Blackwell Publishing Ltd},
ISSN = {1467-8659},
DOI = {10.1111/j.1467-8659.2009.01389.x}
}

@INPROCEEDINGS{TopNet,
  author={Tchapmi, Lyne P. and Kosaraju, Vineet and Rezatofighi, Hamid and Reid, Ian and Savarese, Silvio},
  booktitle={2019 IEEE/CVF Conference on Computer Vision and Pattern Recognition (CVPR)}, 
  title={TopNet: Structural Point Cloud Decoder}, 
  year={2019},
  volume={},
  number={},
  pages={383-392},
  doi={10.1109/CVPR.2019.00047}}

@inproceedings{PoinTr,
author = {Yu, Xumin and Rao, Yongming and Wang, Ziyi and Liu, Zuyan and Lu, Jiwen and Zhou, Jie},
year = {2021},
month = {10},
pages = {12478-12487},
title = {PoinTr: Diverse Point Cloud Completion with Geometry-Aware Transformers},
doi = {10.1109/ICCV48922.2021.01227}
}

@article{SnowflakeNet,
  author       = {Peng Xiang and
                  Xin Wen and
                  Yu{-}Shen Liu and
                  Yan{-}Pei Cao and
                  Pengfei Wan and
                  Wen Zheng and
                  Zhizhong Han},
  title        = {SnowflakeNet: Point Cloud Completion by Snowflake Point Deconvolution
                  with Skip-Transformer},
  journal      = {CoRR},
  volume       = {abs/2108.04444},
  year         = {2021},
  eprinttype   = {arXiv},
  eprint       = {2108.04444},
  bibsource    = {dblp computer science bibliography, https://dblp.org}
}

@article{Variational,
  author       = {Liang Pan and
                  Xinyi Chen and
                  Zhongang Cai and
                  Junzhe Zhang and
                  Haiyu Zhao and
                  Shuai Yi and
                  Ziwei Liu},
  title        = {Variational Relational Point Completion Network},
  journal      = {CoRR},
  volume       = {abs/2104.10154},
  year         = {2021},
  eprinttype   = {arXiv},
  eprint       = {2104.10154},
  bibsource    = {dblp computer science bibliography, https://dblp.org}
}

@article{mao2024dmf,
  title={DMF-Net: Image-Guided Point Cloud Completion with Dual-Channel Modality Fusion and Shape-Aware Upsampling Transformer},
  author={Mao, Aihua and Tang, Yuxuan and Huang, Jiangtao and He, Ying},
  journal={arXiv preprint arXiv:2406.17319},
  year={2024}
}

@article{song2023fine,
  title={Fine-grained text and image guided point cloud completion with clip model},
  author={Song, Wei and Zhou, Jun and Wang, Mingjie and Tan, Hongchen and Li, Nannan and Liu, Xiuping},
  journal={arXiv preprint arXiv:2308.08754},
  year={2023}
}

@article{lin2022cosmos,
  title={Cosmos propagation network: Deep learning model for point cloud completion},
  author={Lin, Fangzhou and Xu, Yajun and Zhang, Ziming and Gao, Chenyang and Yamada, Kazunori D},
  journal={Neurocomputing},
  volume={507},
  pages={221--234},
  year={2022},
  publisher={Elsevier}
}

@article{xu2022fpcc,
  title={FPCC: Fast point cloud clustering-based instance segmentation for industrial bin-picking},
  author={Xu, Yajun and Arai, Shogo and Liu, Diyi and Lin, Fangzhou and Kosuge, Kazuhiro},
  journal={Neurocomputing},
  volume={494},
  pages={255--268},
  year={2022},
  publisher={Elsevier}
}

@inproceedings{lin2024loss,
  title={Loss Distillation via Gradient Matching for Point Cloud Completion with Weighted Chamfer Distance},
  author={Lin, Fangzhou and Liu, Haotian and Zhou, Haoying and Hou, Songlin and Yamada, Kazunori D and Fischer, Gregory S and Li, Yanhua and Zhang, Haichong K and Zhang, Ziming},
  booktitle={2024 IEEE/RSJ International Conference on Intelligent Robots and Systems (IROS)},
  pages={511--518},
  year={2024},
  organization={IEEE}
}

@article{lin2023infocd,
  title={InfoCD: a contrastive chamfer distance loss for point cloud completion},
  author={Lin, Fangzhou and Yue, Yun and Zhang, Ziming and Hou, Songlin and Yamada, Kazunori and Kolachalama, Vijaya and Saligrama, Venkatesh},
  journal={Advances in Neural Information Processing Systems},
  volume={36},
  pages={76960--76973},
  year={2023}
}

@inproceedings{lin2023hyperbolic,
  title={Hyperbolic chamfer distance for point cloud completion},
  author={Lin, Fangzhou and Yue, Yun and Hou, Songlin and Yu, Xuechu and Xu, Yajun and Yamada, Kazunori D and Zhang, Ziming},
  booktitle={Proceedings of the IEEE/CVF international conference on computer vision},
  pages={14595--14606},
  year={2023}
}

@inproceedings{lu2024mmcnet,
  title={MMCNet: Multi-Modal Point Cloud Completion Network Based on Self-Projected Views},
  author={Lu, Ziyuan and Jiao, Qinglong and Xu, Liangdong},
  booktitle={2024 6th International Conference on Data-driven Optimization of Complex Systems (DOCS)},
  pages={897--902},
  year={2024},
  organization={IEEE}
}

@article{zhu2023csdn,
  title={Csdn: Cross-modal shape-transfer dual-refinement network for point cloud completion},
  author={Zhu, Zhe and Nan, Liangliang and Xie, Haoran and Chen, Honghua and Wang, Jun and Wei, Mingqiang and Qin, Jing},
  journal={IEEE Transactions on Visualization and Computer Graphics},
  year={2023},
  publisher={IEEE}
}

@inproceedings{radford2021learning,
  title={Learning transferable visual models from natural language supervision},
  author={Radford, Alec and Kim, Jong Wook and Hallacy, Chris and Ramesh, Aditya and Goh, Gabriel and Agarwal, Sandhini and Sastry, Girish and Askell, Amanda and Mishkin, Pamela and Clark, Jack and others},
  booktitle={International conference on machine learning},
  pages={8748--8763},
  year={2021},
  organization={PmLR}
}

@article{wang2019dynamic,
  title={Dynamic graph CNN for learning on point clouds},
  author={Wang, Yue and Sun, Yongbin and Liu, Ziwei and Sarma, Sanjay E and Bronstein, Michael M and Solomon, Justin M},
  journal={ACM Transactions on Graphics},
  volume={38},
  number={5},
  pages={146:1--146:12},
  year={2019},
  publisher={ACM}
}

@inproceedings{yuan2018pcn,
  title={{PCN}: Point Completion Network},
  author={Yuan, Wentao and Khot, Tejas and Held, David and Mertz, Christoph and Hebert, Martial},
  booktitle={2018 International Conference on 3D Vision (3DV)},
  pages={728--737},
  year={2018},
  organization={IEEE}
}

@inproceedings{yang2018foldingnet,
  title={{FoldingNet}: Point cloud auto-encoder via folding},
  author={Yang, Yaoqing and Feng, Cihui and Shen, Yuwei and Tian, Dong},
  booktitle={Proceedings of the IEEE Conference on Computer Vision and Pattern Recognition},
  pages={206--215},
  year={2018},
  organization={IEEE}
}

@inproceedings{tchapmi2019topnet,
  title={Topnet: Structural point cloud decoder},
  author={Tchapmi, Lyne P and Kosaraju, Vineet and Rezatofighi, Hamid and Reid, Ian and Savarese, Silvio},
  booktitle={Proceedings of the IEEE/CVF conference on computer vision and pattern recognition},
  pages={383--392},
  year={2019}
}

@String(CVPR  = {IEEE Conf. Comput. Vis. Pattern Recog.})

@String(AAAI  = {AAAI})

@String(CVPR  = {CVPR})

@inproceedings{zhang2021view,
  title={View-guided point cloud completion},
  author={Zhang, Xuancheng and Feng, Yutong and Li, Siqi and Zou, Changqing and Wan, Hai and Zhao, Xibin and Guo, Yandong and Gao, Yue},
  booktitle={Proceedings of the IEEE/CVF Conference on Computer Vision and Pattern Recognition},
  pages={15890--15899},
  year={2021}
}

@inproceedings{yu2021pointr,
  title={Pointr: Diverse point cloud completion with geometry-aware transformers},
  author={Yu, Xumin and Rao, Yongming and Wang, Ziyi and Liu, Zuyan and Lu, Jiwen and Zhou, Jie},
  booktitle={Proceedings of the IEEE/CVF international conference on computer vision},
  pages={12498--12507},
  year={2021}
}

@InProceedings{Tchapmi_2019_CVPR,
author = {Tchapmi, Lyne P. and Kosaraju, Vineet and Rezatofighi, Hamid and Reid, Ian and Savarese, Silvio},
title = {TopNet: Structural Point Cloud Decoder},
booktitle = {Proceedings of the IEEE/CVF Conference on Computer Vision and Pattern Recognition (CVPR)},
month = {June},
year = {2019}
}

@InProceedings{Huang_2020_CVPR,
author = {Huang, Zitian and Yu, Yikuan and Xu, Jiawen and Ni, Feng and Le, Xinyi},
title = {PF-Net: Point Fractal Network for 3D Point Cloud Completion},
booktitle = {Proceedings of the IEEE/CVF Conference on Computer Vision and Pattern Recognition (CVPR)},
month = {June},
year = {2020}
}

@inproceedings{zhou2022seedformer,
  title={Seedformer: Patch seeds based point cloud completion with upsample transformer},
  author={Zhou, Haoran and Cao, Yun and Chu, Wenqing and Zhu, Junwei and Lu, Tong and Tai, Ying and Wang, Chengjie},
  booktitle={European conference on computer vision},
  pages={416--432},
  year={2022},
  organization={Springer}
}

@article{qi2017pointnet++,
  title={Pointnet++: Deep hierarchical feature learning on point sets in a metric space},
  author={Qi, Charles Ruizhongtai and Yi, Li and Su, Hao and Guibas, Leonidas J},
  journal={Advances in neural information processing systems},
  volume={30},
  year={2017}
}

@article{yang2017foldingnet,
  title={Foldingnet: Interpretable unsupervised learning on 3d point clouds},
  author={Yang, Yaoqing and Feng, Chen and Shen, Yiru and Tian, Dong},
  journal={arXiv preprint arXiv:1712.07262},
  volume={2},
  number={3},
  pages={5},
  year={2017}
}

@inproceedings{liu2020morphing,
  title={Morphing and sampling network for dense point cloud completion},
  author={Liu, Minghua and Sheng, Lu and Yang, Sheng and Shao, Jing and Hu, Shi-Min},
  booktitle={Proceedings of the AAAI conference on artificial intelligence},
  volume={34},
  number={07},
  pages={11596--11603},
  year={2020}
}

@inproceedings{groueix2018papier,
  title={A papier-m{\^a}ch{\'e} approach to learning 3d surface generation},
  author={Groueix, Thibault and Fisher, Matthew and Kim, Vladimir G and Russell, Bryan C and Aubry, Mathieu},
  booktitle={Proceedings of the IEEE conference on computer vision and pattern recognition},
  pages={216--224},
  year={2018}
}

@inproceedings{xie2020grnet,
  title={Grnet: Gridding residual network for dense point cloud completion},
  author={Xie, Haozhe and Yao, Hongxun and Zhou, Shangchen and Mao, Jiageng and Zhang, Shengping and Sun, Wenxiu},
  booktitle={European Conference on Computer Vision},
  pages={365--381},
  year={2020},
  organization={Springer}
}

@article{wang2022pointattn,
  title={Pointattn: You only need attention for point cloud completion},
  author={Wang, Jun and Cui, Ying and Guo, Dongyan and Li, Junxia and Liu, Qingshan and Shen, Chunhua},
  journal={arXiv preprint arXiv:2203.08485},
  year={2022}
}

@article{zhang2022point,
  title={Point cloud completion via skeleton-detail transformer},
  author={Zhang, Wenxiao and Dong, Zhen and Liu, Jun and Yan, Qingan and Xiao, Chunxia and others},
  journal={IEEE Transactions on Visualization and Computer Graphics},
  year={2022},
  publisher={IEEE}
}

@inproceedings{wu2021densityaware,
  title={Density-aware chamfer distance as a comprehensive metric for point cloud completion},
  author={Wu, Tong and Pan, Liang and Zhang, Junzhe and Wang, Tai and Liu, Ziwei and Lin, Dahua},
  booktitle={Advances in Neural Information Processing Systems},
  volume={34},
  pages={29088--29100},
  year={2021}
}

@inproceedings{xu2024explicitly,
  title={Explicitly guided information interaction network for cross-modal point cloud completion},
  author={Xu, Hang and Long, Chen and Zhang, Wenxiao and Liu, Yuan and Cao, Zhen and Dong, Zhen and Yang, Bisheng},
  booktitle={European Conference on Computer Vision},
  pages={414--432},
  year={2024},
  organization={Springer}
}

@inproceedings{fu2023vapcnet,
  title={VAPCNet: viewpoint-aware 3D point cloud completion},
  author={Fu, Zhiheng and Wang, Longguang and Xu, Lian and Wang, Zhiyong and Laga, Hamid and Guo, Yulan and Boussaid, Farid and Bennamoun, Mohammed},
  booktitle={Proceedings of the IEEE/CVF International Conference on Computer Vision},
  pages={12108--12118},
  year={2023}
}

@article{tang2025calibration,
  title={Calibration between a panoramic LiDAR and a limited field-of-view depth camera},
  author={Tang, Weijie and Wang, Bin and Huang, Longxiang and Yang, Xu and Zhang, Qian and Zhu, Sulei and Ma, Yan},
  journal={Complex \& Intelligent Systems},
  volume={11},
  number={1},
  pages={102},
  year={2025},
  publisher={Springer}
}

@article{zhou2024position,
  title={Position-aware Guided Point Cloud Completion with CLIP Model},
  author={Zhou, Feng and Zhang, Qi and Dai, Ju and Li, Lei and Fan, Qing and Xing, Junliang},
  journal={arXiv preprint arXiv:2412.08271},
  year={2024}
}

@article{hou2025mobile,
  title={Mobile Augmented Reality Framework with Fusional Localization and Pose Estimation},
  author={Hou, Songlin and Lin, Fangzhou and Huang, Yunmei and Peng, Zhe and Xiao, Bin},
  journal={arXiv preprint arXiv:2501.03336},
  year={2025}
}

@article{zhang2021point,
  title={Point set voting for partial point cloud analysis},
  author={Zhang, Junming and Chen, Weijia and Wang, Yuping and Vasudevan, Ram and Johnson-Roberson, Matthew},
  journal={IEEE Robotics and Automation Letters},
  volume={6},
  number={2},
  pages={596--603},
  year={2021},
  publisher={IEEE}
}

@inproceedings{zhang2025gps,
  title={Gps: A probabilistic distributional similarity with gumbel priors for set-to-set matching},
  author={Zhang, Ziming and Lin, Fangzhou and Liu, Haotian and Morales, Jose and Zhang, Haichong and Yamada, Kazunori and Kolachalama, Vijaya B and Saligrama, Venkatesh},
  booktitle={The Thirteenth International Conference on Learning Representations},
  year={2025}
}

@article{zhang2024deep,
  title={Deep Loss Convexification for Learning Iterative Models},
  author={Zhang, Ziming and Shao, Yuping and Zhang, Yiqing and Lin, Fangzhou and Zhang, Haichong and Rundensteiner, Elke},
  journal={IEEE Transactions on Pattern Analysis and Machine Intelligence},
  year={2024},
  publisher={IEEE}
}

@article{gao2025safecoop,
  title={SafeCoop: Unravelling Full Stack Safety in Agentic Collaborative Driving},
  author={Gao, Xiangbo and Lin, Tzu-Hsiang and Song, Ruojing and Wu, Yuheng and Huang, Kuan-Ru and Jin, Zicheng and Lin, Fangzhou and Liu, Shinan and Tu, Zhengzhong},
  journal={arXiv preprint arXiv:2510.18123},
  year={2025}
}

@article{lin2024hyperbolic,
  title={Hyperbolic Chamfer Distance for Point Cloud Completion and Beyond},
  author={Lin, Fangzhou and Hou, Songlin and Liu, Haotian and Gao, Shang and Yamada, Kazunori D and Zhang, Haichong K and Zhang, Ziming},
  journal={arXiv preprint arXiv:2412.17951},
  year={2024}
}

@inproceedings{yue2024understanding,
  title={Understanding hyperbolic metric learning through hard negative sampling},
  author={Yue, Yun and Lin, Fangzhou and Mou, Guanyi and Zhang, Ziming},
  booktitle={Proceedings of the IEEE/CVF Winter Conference on Applications of Computer Vision},
  pages={1891--1903},
  year={2024}
}

\end{document}